\documentclass[10pt,letterpaper]{article}

\usepackage{cvpr}
\usepackage{times}
\usepackage{epsfig}
\usepackage{graphicx}
\usepackage{amsmath}
\usepackage{amssymb}

\usepackage{algorithm}
\usepackage{algpseudocode}
\usepackage{multicol,caption}
\usepackage{multirow}

\usepackage[pagebackref=true,breaklinks=true,letterpaper=true,colorlinks,bookmarks=false]{hyperref}

\newenvironment{Figure}
  {\par\medskip\noindent\minipage{\linewidth}}
  {\endminipage\par\medskip}

 \cvprfinalcopy 

\def\cvprPaperID{0977} 

\ifcvprfinal\fi
\begin{document}

\title{A Fast Two Pass Multi-Value Segmentation Algorithm based on Connected Component Analysis}

\author{Dibyendu Mukherjee\\
University of Windsor\\
401 Sunset Avenue, Windsor, Canada\\
{\tt\small dibyendu.mukherjee@ieee.org}
}

\maketitle
\begin{multicols}{2}
\begin{abstract}
   Connected component analysis (CCA) has been heavily used to label binary images and classify segments. However, it has not been well-exploited to segment multi-valued natural images. This work proposes a novel multi-value segmentation algorithm that utilizes CCA to segment color images. A user defined distance measure is incorporated in the proposed modified CCA to identify and segment similar image regions. The raw output of the algorithm consists of distinctly labelled segmented regions. The proposed algorithm has a unique design architecture that provides several benefits: 1) it can be used to segment any multi-channel multi-valued image; 2) the distance measure/segmentation criteria can be application-specific and 3) an absolute linear-time implementation allows easy extension for real-time video segmentation. Experimental demonstrations of the aforesaid benefits are presented along with the comparison results on multiple datasets with current benchmark algorithms. A number of possible application areas are also identified and results on real-time video segmentation has been presented to show the promise of the proposed method.
\end{abstract}

\section{Introduction}
\label{sec:intro}

Connected components analysis is a well-explored fundamental topic in image \& video processing. In binary images, CCA is used to segment 4 or 8-connected regions by assigning unique labels. The classical approaches for CCA date back to the 1960s~\cite{rosenfeld:book,rosenfeld:1966} and mainly focussed on labeling binary images. These labels can be used for shape characterizations to object recognition. Even today, binary CCA is very popular for object detection, tracking, segmentation and other computer vision applications. With the diversity of applications, methods for CCA have also improved over time~\cite{grana:CCA,lifeng:CCA,suzuki:CCA,wu:CCA}. An overview of the advancements is provided in \cite{grana:CCA}. Although the improvements in the area of CCA is worth noticing, very few of the\begin{Figure}
\begin{center}
\includegraphics[width=\linewidth]{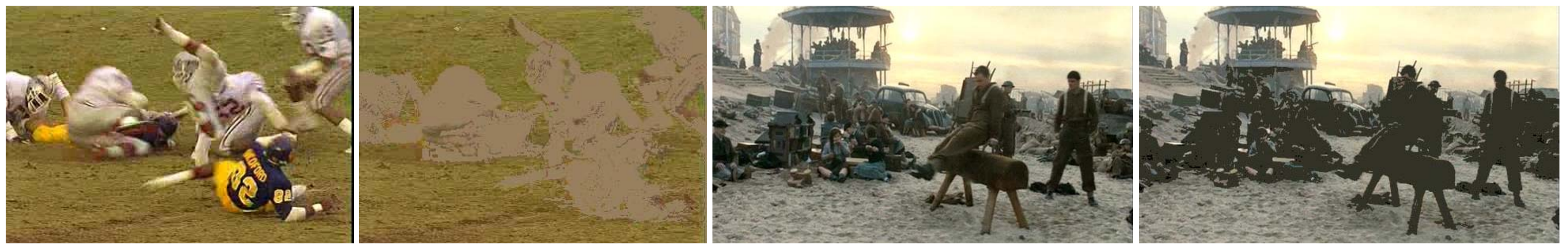}
\end{center}
\vspace{-0.1in}
\captionof{figure}{\small Conditional segmentation results of the proposed method.}
\label{fig:titleImage}
\end{Figure}\hspace{-0.17in}approaches had their focus on labeling connected components in color images. Binary images consist of two classes of values: 0 and 1. It is simpler to treat the pixels to be part of a region (value of 1) or no region (value of 0) and segment accordingly. However, for color images, the problem of labeling is more complicated due to the followings: 1) each pixel belongs to a region; 2) the regions are not unique in terms of color. As color values are randomly distributed in a natural image, it is not possible to classify a region in terms of a single color value, without specific criteria. There have been a few approaches for color image segmentation based on CCA\cite{celebi:colorSegCCA,mandler:colorSegCCA,Maresca:colorSegCCA,sang:colorSegCCA}. However, most of them are limited to constant color regions and higher computational complexity. This work proposes a linear-time two-pass CCA based multi-valued image segmentation algorithm, and provides different measures for classification of a color region in terms of color values, Gradient values and Saliency values. The algorithm is based on similarity of neighboring pixels. The architecture is influenced from the two-pass binary CCA proposed by Haralick~\cite{haralick:book}. Haralick's algorithm has a space efficient run-length implementation. This implementation has been partially adopted with significant changes: : 1) The algorithm has been introduced for multi-valued image data; 2) unlike Haralick's algorithm, the initialization of data structures are carried out in the top-down pass. So, there are no hidden computational costs involved; 3) the algorithm has a modular architecture in which, the segmentation criteria is called from outside the original function flow signifying a completely user-specific criteria that can be chosen by a run-time polymorphism in any advanced computer language that supports it; 4) The algorithm does not require the number of clusters as input, and 5) the implementation renders it suitable for real-time applications or streaming videos.

The algorithm is explained in Section~\ref{sec:algo}, followed by a discussion on the distance measures for segmentation~\ref{sec:distMeasure}. The experimental results are reported in Section~\ref{sec:exp}. Finally, the paper is concluded in Section~\ref{sec:conclusion}.

\section{Algorithm}
\label{sec:algo}

A number of important definitions are required before the explanation of the proposed algorithm. The definitions are provided with respect to an image $IM$ of $N_R$ rows, $N_C$ columns and $N_{CH}$ number of color channels:

\begin{description}\small
\item[Similarity:] Two pixels $PX_p$ and $PX_q$ at locations $(r_p,c_p)$ and $(r_q,c_q)$ respectively, are termed \emph{similar} if they share  a \emph{common property}. This \emph{common property} can be defined in terms of color similarity, gradient orientation similarity or any other similarity or distance measurement operation, depending on the application.
\item[Run:] A run is defined by a set of contiguous and \emph{similar} pixels, in a particular row. It is uniquely represented by the row index, and its starting and ending column index.
\item[Equivalence:] Two runs $p$ and $q$ are equivalent if they satisfy at least one of the following criteria:
    \begin{enumerate}
        \item TYPE-I: $p$ and $q$ share a common boundary and any two pixels $PX_p$ and $PX_q$ associated with $p$ and $q$ respectively, are \emph{similar} using the \emph{common property} criteria.
        \item TYPE-II: $p$ and $q$ are equivalent to runs $u$ and $v$ respectively, using criteria 1. If $u$ and $v$ are proved equivalent, $p$ and $q$ are also equivalent.
    \end{enumerate}
\item[Label:] Apart from its index, a run also has a permanent label. When two runs are equivalent, they are merged by assigning them a single label. In a segmented image, all equivalent runs are assigned same labels.
\end{description}

In this work, \emph{similarity} is verified using a function $dist(r_p,c_p,r_q,c_q)$ that takes pixel coordinate pairs $(r_p,c_p)$ and $(r_q,c_q)$ as input, and returns true, if $PX_p$ and $PX_q$ are \emph{similar}. Otherwise, it returns false. There can be several runs in a row. A run ends when its end pixel is not \emph{similar} to its subsequent pixel in the same row, or if the row ends. An example image is shown in Figure~\ref{fig:Datastructures}. The first row has first 4 elements of red color, the next single element of blue color, followed by 3 more elements in red and the last two elements in green. Hence, there are four runs in the row: red (1-4), blue (5-5), red (6-8) and green (9-10). Considering $dist()$ to have a threshold (typically 10-15) on Euclidean distance between two color pixels for this example, all ``reddish" pixels are considered as ``Red" and similar for Blue and Green. Thus, second row has 6 runs while third row has 3 runs.

The algorithm has two passes. In the top-down pass, the runs are computed, labels are assigned, and equivalences are established. In the bottom-up pass, equivalent runs are assigned same labels. Hence, the image is segmented. In the following paragraphs, the two passes are discussed in details with related data structures. Finally, the pseudo-codes of the algorithm are presented using three procedures: \textsc{Segment}~\ref{Segment}, \textsc{InitLabel}~\ref{InitLabel}, \textsc{MakeEquivalent}~\ref{MakeEquivalent}.

In the top-down pass, the runs are computed and their information are populated in two data structures: $rowBlock$ and $colBlock$. $rowBlock$ has a dimension of $N_R \times 2$ and keeps the number of runs present in each row. The two elements $firstRun$ and $lastRun$ of $i^{\textrm{th}}$ row in $rowBlock$ keep the index of the first run and last run in $i^{\textrm{th}}$ row of $IM$, respectively. In the example image, the first row has four runs indexed 1 to 4. Thus, $rowBlock(1).firstRun = 1, rowBlock(1).lastRun = 4$. Here, $rowBlock(i)$ represent the $i^\textrm{th}$ row, and the ``$\cdot$" operator is used to refer to the elements of that row.

\begin{Figure}
\begin{center}
\includegraphics[width=\linewidth]{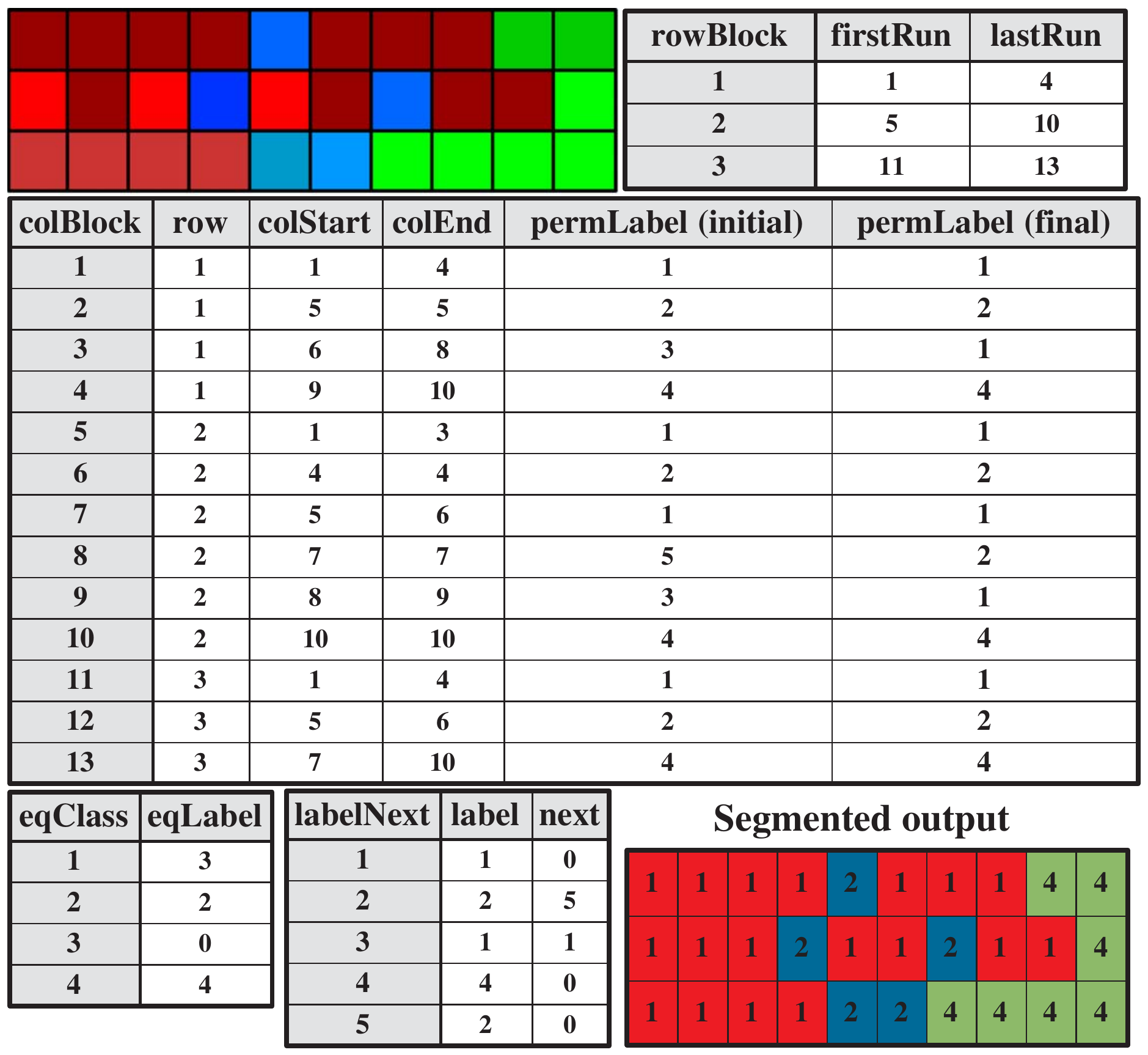}
\end{center}
\vspace{-0.1in}
\captionof{figure}{\small The data structures for a sample run on an image. From top-left to bottom-right: the image, $rowBlock$, $colBlock$, $eqClass$, $labelNext$ and segmented output.}
\label{fig:Datastructures}
\end{Figure}

$colBlock$ has the dimension of $N_{RN} \times N_C + 4$ with $N_{RN}$ representing the number of runs. $i^{\textrm{th}}$ row represents the necessary information for $i^{\textrm{th}}$ run. The elements $row$, $colStart$, $colEnd$ and $permLabel$ in $i^{\textrm{th}}$ row of $colBlock$, represent the row index, start and end column index, and permanent label for the $i^{\textrm{th}}$ run. In the top-down pass for $p^{\textrm{th}}$ run in $r^{\textrm{th}}$ row, every pixel $PX_p$ is compared with its three neighboring pixels in $(r-1)^{\textrm{th}}$ row (if the row exists), using the $dist()$ function. If a neighboring pixel $PX_q$ belonging to run $q$ is similar to $PX_p$, the runs $p$ and $q$ are TYPE-I equivalent. Thus, they are made equivalent using a procedure \textsc{MakeEquivalent}~\ref{MakeEquivalent}. Finally, the $permLabels$ are replaced with the equivalent labels in the bottom-up pass as provided in procedure \textsc{Segment}~\ref{Segment}. An image and its sample run information are provided in Figure~\ref{fig:Datastructures}. The initial labels after top-down pass, and final labels after bottom-up pass are provided in $colBlock$ as well.

Finally, TYPE-II equivalences are resolved using the two data structures: $eqClass$ and $labelNext$. If runs $p$ and $q$ with $permlabels$ $L_p$ and $L_q$ respectively, are equivalent, they must belong to same \emph{class} $c$. A \emph{class} $c$ has a \emph{unique} label $L_c$, and the $permLabels$ of $p$ and $q$ are made equivalent by associating both of them to $L_c$. The $labelNext$ structure has two elements: $label$ and $next$. If $p$ already belongs to \emph{class} $c$, $labelNext(L_p).label = L_c (\neq 0)$. If $p$ and $q$ need to be made equivalent, $L_q$ needs to be associated to $L_c$. If $q$ does not belong to any \emph{class} \ie $labelNext(L_q).label = 0)$, the procedure is easy. However, if $q$ belongs to \emph{class} $\hat{c}$ \ie $labelNext(L_q).label = L_{\hat{c}} (\neq 0 \neq L_c)$, $c$ and $\hat{c}$ should be equivalent. This equivalence is kept in data structure $eqClass$ element $eqLabel$. Thus, in this case, $labelNext(L_q).label$ is made equal to $L_c$ and $eqClass(L_c)$ is made equal to $L_{\hat{c}}$ to link both the classes. This process is established through the procedure \textsc{MakeEquivalent}~\ref{MakeEquivalent} depicted in the algorithm.

\begin{algorithm*}\scriptsize
\caption{\small Two-Pass Segmentation Algorithm}
\label{Segment}
\begin{multicols}{1}
\begin{algorithmic}[1]
\Procedure{Segment}{}
\State $cr = 0;$
\State $cl = 0;$
\For{$r = 1, N_R$}\Comment{Top-down pass}
    \State $fr = \textrm{false};$
    \State $cx = -1; cy = -1;$
    \For{$c = 1, N_C$}
        \State $matched = \textrm{dist}(r, c, cx, cy)$;
        \If{$(fr)$}
            \If{$matched$}
                \State $cx = c; cy = r;$
            \Else
                \State $fr = \textrm{false};$
            \EndIf
        \EndIf
        \If{$(fr = \textrm{false}) \& (matched = \textrm{false})$}
            \If{$(cr>0)$}
                \If{$colBlock(cr).label == 0$}
                    \State $cl = cl + 1;$
                    \State $colBlock(cr).permLabel = cl;$
                \EndIf
            \EndIf
            \State $cr = cr + 1;$
            \State $cx = c; cy = r;$
            \State $colBlock(cr).row = r;$
            \State $colBlock(cr).colStart = c;$
            \State $colBlock(cr).permLabel = 0;$
            \State $fr = \textrm{true};$
            \If{$rowBlock(r).firstRun == 0$}
                \State $rowBlock(r).firstRun = cr;$
            \EndIf
            \State $rowBlock(r).lastRun = cr;$
        \EndIf
        \If{$(r>1)$}
            \State $\textsc{InitLabel}(r,c);$
        \EndIf
        \If{$fr$}
            \State $colBlock(cr).colEnd = c;$
        \EndIf
        \State $idxImg(r,c) = cr;$
    \EndFor
\EndFor

\If{$(colBlock(cr).permLabel)==0$}
    \State $cLabel = cLabel + 1;$
    \State $colBlock(cr).permLabel = cLabel;$
\EndIf

\For{$r = imSize(1), -1, 1$}\Comment{Bottom-up pass}
    \State $p = rowBlock(r).firstRun;$
    \State $pLast = rowBlock(r).lastRun;$ 
    \If{$(p \neq 0)$}
        \While{$(p<=pLast)$}
            \State $pl = colBlock(p).permLabel;$ 
            \State $ql = labelNext(pl).label$;
            \If{$ql  \neq  0$}
                \State $colBlock(p).permLabel = ql;$
            \EndIf
            \State $p = p + 1;$
        \EndWhile
    \EndIf
\EndFor
\EndProcedure
\end{algorithmic}
\end{multicols}
\vspace{-0.1in}
\end{algorithm*}

\begin{algorithm*}\scriptsize
\caption{\small Initial Labeling}
\label{InitLabel}
\begin{multicols}{1}
\begin{algorithmic}[1]
\Procedure{InitLabel}{$r,c,cr$}
\State $matched = \textrm{dist}(r, c, r - 1, c);$
\If{$matched$}
    \State $tr = idxImg(r-1,c);$
    \State $pl = colBlock(cr).permLabel;$
    \State $ql = colBlock(tr).permLabel;$
    \If{$pl == 0$}
        \State $colBlock(cr).permLabel = ql;$
    \Else
        \State $\textsc{MakeEquivalent}(pl,ql);$
    \EndIf
\EndIf
\If{$(c>1)$}
    \State $matched = \textrm{dist}(r, c, r - 1, c - 1);$
    \If{$matched$}
        \State $tr = idxImg(r-1,c-1);$
        \State $pl = colBlock(cr).permLabel;$
        \State $ql = colBlock(tr).permLabel;$
        \If{$pl == 0$}
            \State $colBlock(cr).permLabel = ql;$
        \Else
            \State $\textsc{MakeEquivalent}(pl,ql);$
        \EndIf
    \EndIf
\EndIf
\If{$(c<imSize(2))$}
    \State $matched = \textrm{dist}(r, c, r - 1, c + 1);$
    \If{$matched$}
        \State $tr = idxImg(r-1,c+1);$
        \State $pl = colBlock(cr).permLabel;$
        \State $ql = colBlock(tr).permLabel;$
        \If{$pl == 0$}
            \State $colBlock(cr).permLabel = ql;$
        \Else
            \State $\textsc{MakeEquivalent}(pl,ql);$
        \EndIf
    \EndIf
\EndIf
\EndProcedure
\end{algorithmic}
\end{multicols}
\vspace{-0.1in}
\end{algorithm*}

\begin{algorithm*}\scriptsize
\caption{\small Make Two Labels Equivalent}
\label{MakeEquivalent}
\begin{multicols}{1}
\begin{algorithmic}[1]
\Procedure{MakeEquivalent}{$pl,ql$}
\State $lpl = labelNext(pl).label;$
\State $lql = labelNext(ql).label;$
\If{$(lpl==0) \& (lql==0)$}
    \State $labelNext(pl).label = pl;$
    \State $labelNext(ql).label = pl;$
    \State $labelNext(pl).next = ql;$
    \State $labelNext(ql).next = 0;$
    \State $eqClass(pl).eqLabel = pl;$
\ElsIf{$lpl == lql$}
    \State Do nothing
\ElsIf{$(lpl \neq 0) \& (lql==0)$}
    \State $bgn = lpl;$
    \State $label(ql).label = bgn;$
    \State $label(ql).next = eqClass(bgn).eqLabel;$
    \State $eqClass(bgn).eqLabel = ql;$
\ElsIf{$(lql \neq 0) \& (lpl==0)$}
    \State $bgn = lql;$
    \State $label(pl).label = bgn;$
    \State $label(pl).next = eqClass(bgn).eqLabel;$
    \State $eqClass(bgn).eqLabel = pl;$
\ElsIf{$(lql \neq 0) \& (lpl \neq 0)$}
    \State $bgn = lql;$
    \State $member = eqClass(bgn).eqLabel;$
    \State $eql = lpl;$
    \While{$label(member).next \neq 0$}
        \State $label(member).label = eql;$
        \State $member = label(member).next;$
    \EndWhile
    \State $label(member).label = eql;$
    \State $label(member).next = eqClass(eql).eqLabel;$
    \State $eqClass(eql).eqLabel = eqClass(bgn).eqLabel;$
    \State $eqClass(bgn).eqLabel = 0;$
\EndIf
\EndProcedure
\end{algorithmic}
\end{multicols}
\vspace{-0.1in}
\end{algorithm*}


\section{The Distance Measure}
\label{sec:distMeasure}

The distance measure is used to identify TYPE-I equivalence of two runs based on neighboring pixel relationships, and directly controls the quality of segmentation. Based on the relationships between two neighboring pixels, they can be part of same segment or different segment. A bare format of the function is as follows:
\begin{algorithmic}[1]\scriptsize
\Function{$dist$}{$r_1,c_1,r_2,c_2$}
\State $similar = \textrm{false};$
\If{$r_2>0$}
    \State $similar = \langle \textrm{custom logic to validate similarity} \rangle$
\EndIf
\State \textsc{Return} {$similar$}
\EndFunction
\end{algorithmic}

In the example Figure~\ref{fig:Datastructures}, an example distance measure is as follows:
\begin{equation}
\label{eq:dist}
similar = \| IM(r_1,c_1)-IM(r_2,c_2) \| <Th,
\end{equation}
Where, $IM(R_1,c_1)$ represent the color information of pixel $(r_1,c_1)$. $\|\cdot\|$ represents the Euclidean norm, and $Th$ is a distance threshold, typically ranging from $0 - 50$ for a pixel value range of $0 - 255$. Thus, $similar$ is true for pixels with a small color difference. For Figure~\ref{fig:Datastructures}, $Th$ normally lies in the range of $5-10$. With $Th = 0$, the pixels in a neighborhood sharing equal color values would be segmented in a distinct region.

Based on this idea, there may be a large number of possible distance measures. If only reddish pixels are needed to be put together, both pixels can be compared to Red and a joint threshold can be put. Similarly, edge information can be included to segment regions separated by edges. However, due to limited space, a discussion on only three types of distance measure is provided. The first one is the Euclidean distance already mentioned in Eq.~\ref{eq:dist}. The other two types discussed are: Gradient based distance measure, and Saliency based distance measure.

\subsection{Gradient based Distance Measure}
\label{subsec:gradientDist}

Gradient $I_G$ of an image is the second norm of the partial differentials $I_X$ and $I_Y$ of the image. Partial differentials can be computed by convolving the image with differential filters along $X$ and $Y$ directions, \eg Sobel, Prewitt etc. Qualitatively, $I_G$ provides the edge information of the image and it contains the magnitude of the differentiation. In this work, simple differential filters, two basic differential filters $H_X = [-1,0,1]$ and $H_Y = {H_X}^T$ are used to provide the differentials along $X$ and $Y$ directions, respectively. Here, $T$ denotes the transpose. Mathematically,
\begin{equation}
\label{eq:updateConventional}
I_X = I \ast H_X; I_Y = I \ast H_Y; I_G = \sqrt{{I_X}^2 + {I_Y}^2}.\\
\end{equation}

Here, $\ast$ denotes convolution. The Gradient information can be used many ways to make a distance measure. However, in this work, individual thresholdings are used:
\begin{equation}
\label{eq:gradientDist}
similar = (I_G(r_1,c_1)<Th) \& (I_G(r_2,c_2)<Th).
\end{equation}

Equation~\ref{eq:gradientDist} is based on the nature of Gradient information. $I_G$ has low value in relatively uniform regions while having a high value in boundary areas. Two pixels are not \emph{similar} if at least one of them reside on edges having a high Gradient value. The difference between simple Euclidean distance based segmentation and Gradient information based segmentation has been demonstrated in the experiments Section~\ref{sec:exp}.

\subsection{Saliency based Distance Measure}
\label{saliencyDist}

Saliency is based on Human Visual System (HVS) and tries to provide a contrast between regions based on their importance to HVS. A foreground having distinctive features can be more salient to the eyes of an observer compared to a relatively uniform background. There have been many algorithms to find salient regions of an image. Considering the execution performance of the methods, the work by Achanta \etal~\cite{achantaSaliency} is found to be well-suited for this work. To reduce duplication, the theory of the method is not presented in this work. Interested readers are encouraged to go through the reference for details. For now, it would suffice to say that the method takes an RGB color image as input, converts it to CIE Lab space, and provides the saliency map $S$ containing Saliency value of each pixel as the Euclidean distance between the pixel's Lab value and the mean Lab.

The distance measure can be derived in a number of ways depending on the application and implementation. However, the main focus of this example is to show how the saliency information can be used to segment the salient regions or the non-salient regions of the image. To properly segment the salient regions, the distance measure is as follows:
\begin{algorithmic}[1]\scriptsize
\State $Sm = \textrm{mean}(S)$;
\If{$(S(r_1,c_1)<Sm) \& (S(r_2,c_2)<Sm)$}
    \State $similar = \| IM(r_1,c_1)-IM(r_2,c_2) \| < Th2$
\Else
    \State $similar = \| IM(r_1,c_1)-IM(r_2,c_2) \| < Th$
\EndIf
\end{algorithmic}

Here, $Sm$ denotes the mean of the Saliency image $S$. $Th2$ is a low threshold typically $[{1/5} - {1/10}]^{\textrm{th}}$ of $Th$. The conditional statements have a purpose. Generally, each segment is treated as a separate \emph{object}. If both pixels have Saliency below a margin ($Sm$), they belong to non-salient objects and need less attention for segmentation. A low threshold $Th2$ is used as number of segments or objects are less important. If otherwise, at least one pixel is salient, a higher threshold $Th$ is used to loosen the constraint of color when segmenting, so that if the pixels belong to same salient object but have distant colors, they can still be merged into one object.

To segment non-salient regions instead, the condition in line 3 can simply be changed as: $(S(r_1,c_1)>Sm) \textrm{OR} (S(r_2,c_2)>Sm)$. The effect of changing the condition is shown in Figure~\ref{fig:saliencyResults}. In the middle image, the salient region is highly segmented using condition in line 3. But, in the right-most image, with a reverse condition, salient region has more textures that are not segmented. More results are provided in Section~\ref{sec:exp}.

\begin{Figure}
\begin{center}
\includegraphics[width=\linewidth]{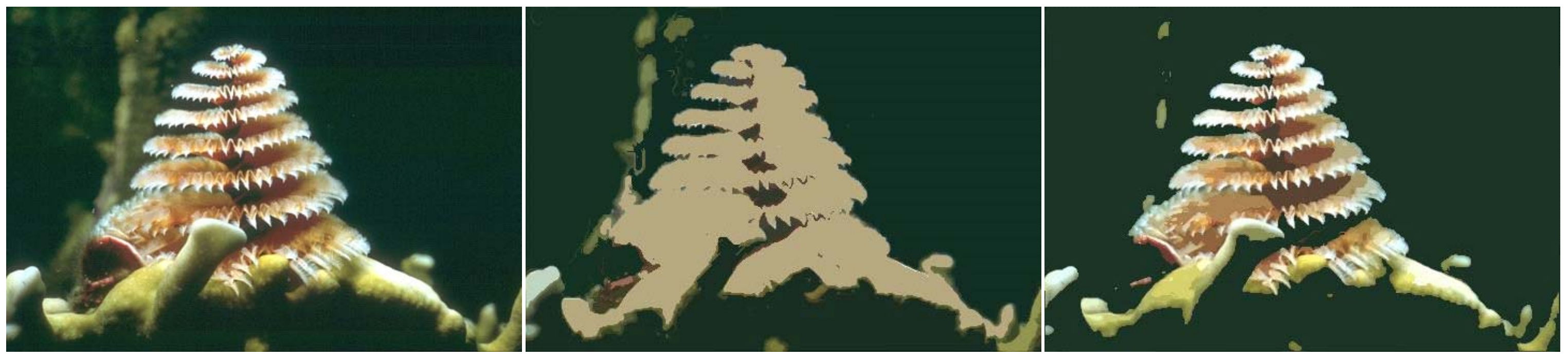}
\end{center}
\vspace{-0.1in}
\captionof{figure}{\small Saliency effect: left - original image, middle - salient segmentation, right - nonsalient segmentation}
\label{fig:saliencyResults}
\end{Figure}

\section{Experimental Results}
\label{sec:exp}

\begin{figure*}
\begin{center}
\includegraphics[width=0.9\linewidth]{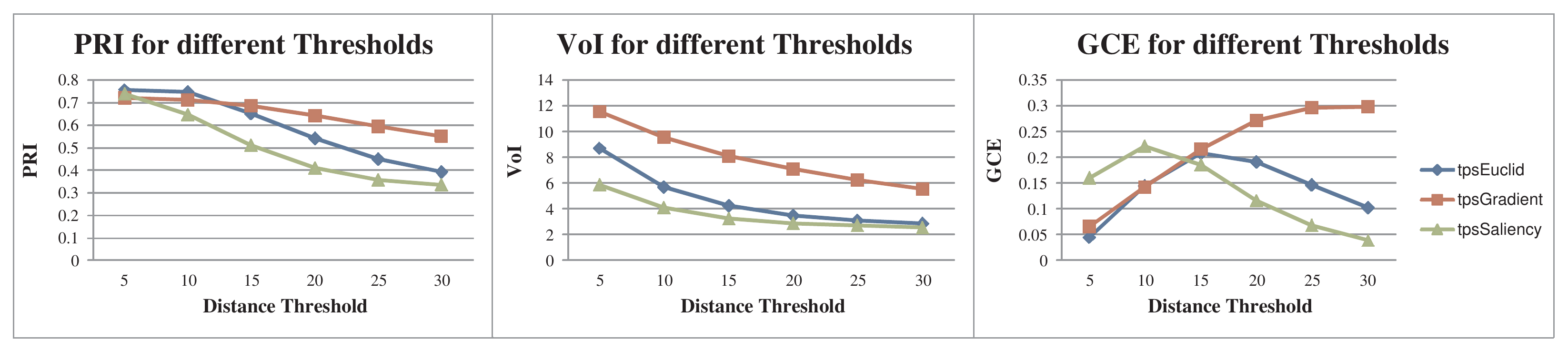}
\end{center}
\vspace{-0.1in}
\caption{\small Variation of PRI, VoI and GCE with variation in distance thresholds.}
\label{fig:graphs}
\end{figure*}

\begin{figure*}
\begin{center}
\includegraphics[width=0.8\linewidth]{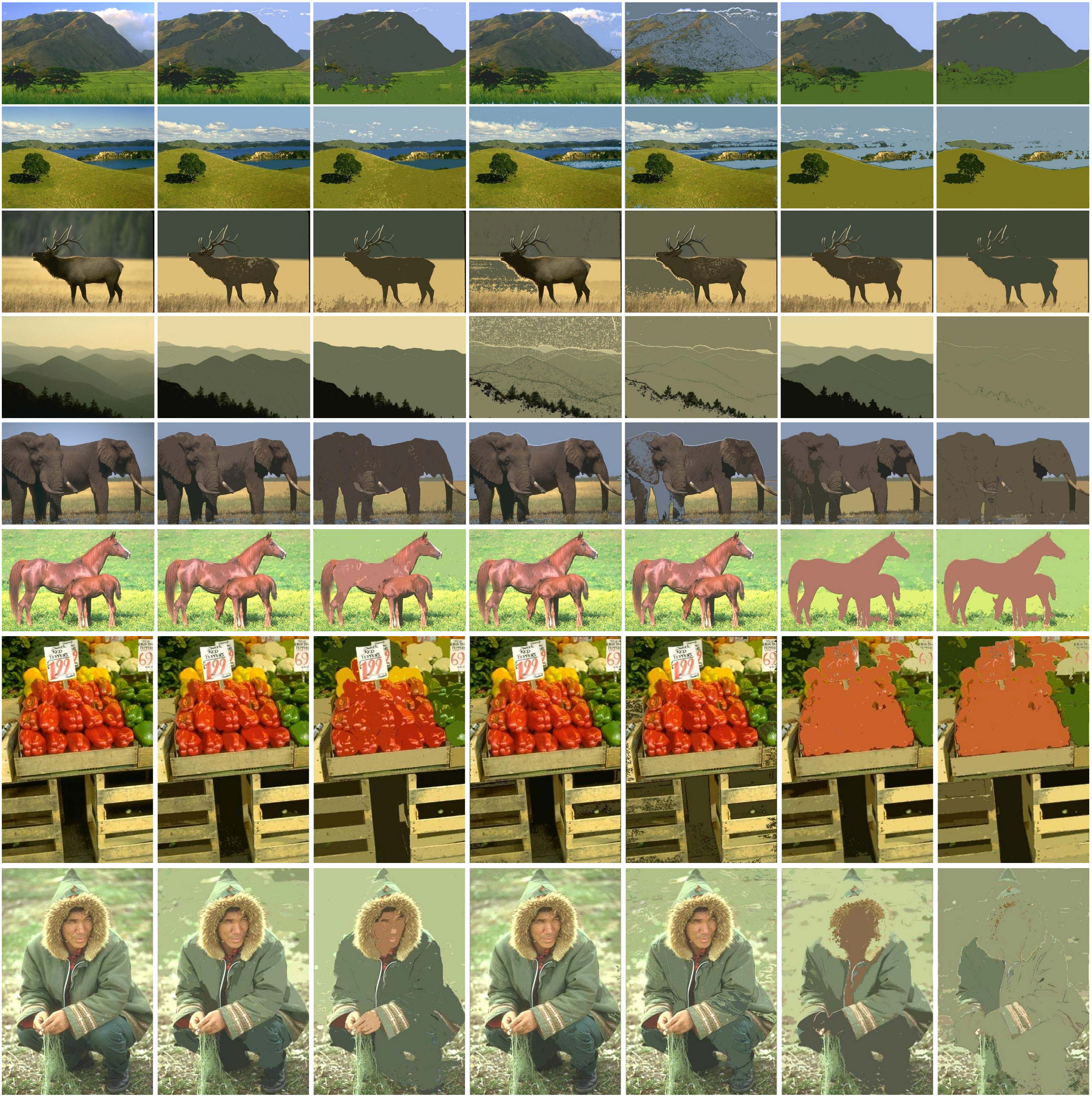}
\end{center}
\vspace{-0.1in}
\caption{\small Qualitative results: First column - original image, second and third columns represent tpsEuclid results with distance threshold 5 and 10, respectively. Similarly, fourth and fifth columns represent tpsGradient and sixth and seventh columns represent tpsSaliency for thresholds 5 and 10, respectively. }
\label{fig:qualitativeResults}
\end{figure*}

The algorithm has been coded in Java and a GUI has been made for ease of use. Java has been chosen mainly because, it has the capability of run-time polymorphism while having a fast execution like C. Using polymorphism, we can define as many definitions of $dist()$ as we need, and choose to call whichever required in run-time. In this section, the three distance measures defined in Section~\ref{sec:distMeasure} have been used and several types of results are provided. To properly address the proposed method with different distance measures, the following nomenclature has been followed: Two pass segmentation method (tps) with 1) Euclidean distance measure - tpsEuclid, 2) Gradient distance measure - tpsGradient and 3) Saliency based measure - tpsSaliency. Image segmentation results are reported on the Berkeley Segmentation databases: BSDS300~\cite{BSDS300} (with 300 images) and BSDS500~\cite{BSDS500} (with 500 images) consisting of natural images from various scene categories. Each image has been segmented by several human subjects. Thus, there exist multiple groundtruths for each image.

Both qualitative and quantitative results are reported in this work. For quantitative results, we used the common criteria used to compute segmentation errors: 1) Probabilistic Rand Index (PRI) - measures the likelihood of a pixel pair being grouped consistently in two segmentations, 2) Variation of Information (VoI) - computes the amount of information in one result not part of the other one, and 3) Global Consistency Error (GCE) - measures the extent to which one segmentation is a refinement of the other one. When compared to groundtruth, a higher value for PRI and lower values for VoI and GCE denote better results. As there are multiple groundtruths, the results are averaged over all the groundtruths for each image.

The results are divided into separate subsections according to their objectives. Section~\ref{sec:distComp} provides a qualitative and quantitative comparison among the three proposed distance measures on BSDS500 with different threshold values. Section~\ref{sec:quali&QuantResults} is dedicated to quantitative comparison of the proposed method with some of the benchmark algorithms from current literatures. An estimation of execution time has been provided in Section~\ref{sec:execTime}. Finally, possible applications are discussed in Section~\ref{sec:vidSeg}.

\subsection{Comparison Among Different Distance Measures}
\label{sec:distComp}

Different distance measures produce distinct segmentations on same image. However, the quality of segmentation also varies greatly with the distance threshold used. In this section, the three methods tpsEuclid, tpsGradient and tpsSaliency have been run on the 500 images of BSDS500 for distance thresholds 5, 10, 15, 20, 25 and 30. The graph in Figure~\ref{fig:graphs} presents the variation of PRI with the variation of distance threshold used for each method. With increase of threshold, distinctly different segments are progressively merged yielding lower quality for PR. However, lower amount of segments also reduce VoI. GCE, on the other hand, depends more on the mapping of segments rather than number of segments, and follows a separate trend.

Some qualitative results are shown in Figure~\ref{fig:qualitativeResults}. According to the images, for different image content, different distances are more suitable. Specifically, images with high salient regions can be better segmented by tpsSaliency, whereas images with higher number of distinct segments are better suited for tpsEuclid and tpsGradient.

\subsection{Benchmark Comparison Results}
\label{sec:quali&QuantResults}

In this section, quantitative comparisons are carried out with some of the benchmark algorithms from current literature. The following algorithms are tested: Ncut~\cite{nCuts}, MShift~\cite{mShift}, FH~\cite{fh}, JSEG~\cite{jSeg}, Multi-scale Ncut (MNcut)~\cite{mnCut}, Normalized Tree Partitioning (NTP)~\cite{ntp}, Saliency Driven Total Variation (SDTV)~\cite{sdtv}, Texture and Boundary Encoding-based Segmentation (TBES)~\cite{tbes} and Segmentation by Aggregating Superpixels (SAS)~\cite{sas}. The scores of the algorithms are collected from \cite{sas}.


\begin{Figure}\scriptsize
\begin{center}
  \captionof{table}{\small Quantitative comparison of proposed method with other methods}
  \begin{tabular}{|c|c|c|c|}
  \hline
  Methods & PRI & VoI & GCE\\
  \hline
  \hline
  Ncut~\cite{nCuts} & 0.7242 & 2.9061 & 0.2232\\
  MShift~\cite{mShift} & 0.7958 & 1.9725 & 0.1888\\
  FH~\cite{fh} & 0.7139 & 3.3949 & 0.1746\\
  JSEG~\cite{jSeg} & 0.7756 & 2.3217 & 0.1989\\
  MNcut~\cite{mnCut} & 0.7559 & 2.4701 & 0.1925\\
  NTP~\cite{ntp} & 0.7521 & 2.4954 & 0.2373\\
  SDTV~\cite{sdtv} & 0.7758 & 1.8165 & 0.1768\\
  TBES~\cite{tbes} & 0.80 & 1.76 & N/A\\
  SAS~\cite{sas} & 0.8319 & 1.6849 & 0.1779\\
  tpsEuclid & 0.7602 & 8.6167 & 0.0446\\
  tpsGradient & 0.7129 & 11.5739 & 0.0691\\
  tpsSaliency & 0.7359 & 5.8860 & 0.1576\\
  \hline
  \end{tabular}
  \label{table:1}
\end{center}
\end{Figure}

The quality of results entirely depends on $dist()$ and may improve with other distance measures. Also, the application areas of the proposed approach are not limited to image segmentation due to its uniqueness of cluster independence, application-specific distance measure and real-time applicability. Regarding real-time applicability, the proposed method is much faster compared to any other algorithm (reported rates of the benchmark algorithms are at least more than 5 seconds/image of size $481 \times 321$). An estimation of time complexity has been provided in the next section.

\subsection{Execution Time Estimation}
\label{sec:execTime}

In this section, average execution times are tabulated for different image sizes. The execution time of the two-pass algorithm depends on the number of runs. With lower value of runs, the passes over each row takes lower time as each pass loops over each run in a row. Thus, the maximum number of iteration arises when, number of runs equals the number of pixels and the algorithm iterates over each pixel twice (top-down and bottom-up) achieving an absolute \texttt{O}(n) complexity. To achieve this extreme condition, a $dist()$ function is used to always return false. As this does not depend on image content any more, an image with random pixel values is used. The original image size is $4096 \times 4096 \times 3$, and it is reduced dyadically in size to as small as $128 \times 128 \times 3$. The times taken by the two-pass algorithm are tabulated in Table~\ref{table:2} in milliseconds (second \& third image dimensions are not shown). However, this does not include the time taken to compute the mean of each segment and display the image on a computer screen or save it, as these are not part of the original algorithm. For the execution, a desktop computer with 3 GHz AMD Phenom II X6 Processor is used.

Referring to the table, if the execution times are plotted on a graph against the numbers of pixels, it will be an approximation of a straight line. This signifies the linear time complexity of the algorithm.

\begin{Figure}\scriptsize
\begin{center}
  \captionof{table}{\small Execution Time Estimation}
  \begin{tabular}{|c||c|c|c|c|c|c|}
  \hline
  Size & $4096$ & $2048$ & $1024$ & $512$ & $256$ & $\leq 128$\\
  \hline
  Time (ms) & $755$ & $70$ & $16$ & $5$ & $1$ & $<1$\\
  \hline
  \end{tabular}
  \label{table:2}
\end{center}
\end{Figure}


\subsection{Applications}
\label{sec:vidSeg}

\begin{Figure}
\begin{center}
\includegraphics[width=\linewidth]{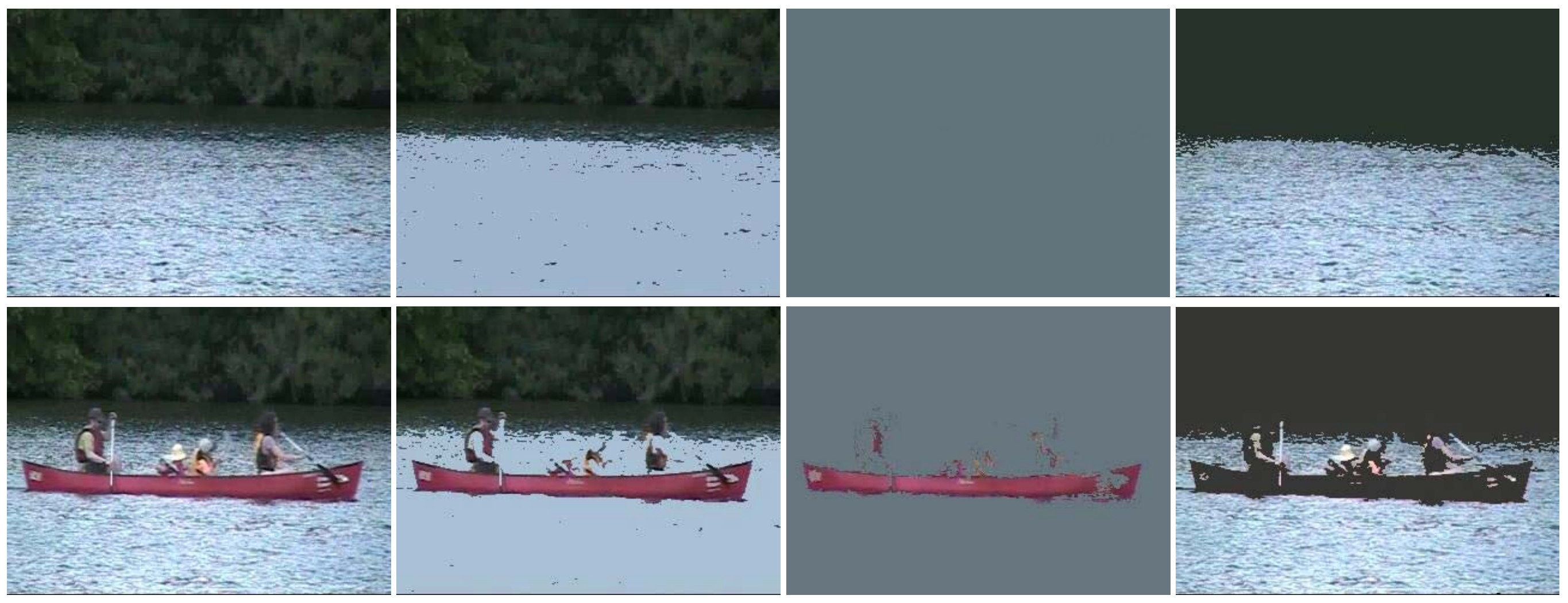}
\end{center}
\vspace{-0.1in}
\captionof{figure}{\small Detection and tracking: First, second, third \& fourth columns shows the original frames, segmented water, segmented boat, and segmented non-water regions, respectively.}
\label{fig:vidSegResults1}
\end{Figure}

The proposed method has several applications. It can be used as a preprocessing step for any image or video based algorithm like conditional segmentation (\eg Figure~\ref{fig:titleImage}), motion or event detection, and tracking. One of the main advantages of using the algorithm is that the outputs contain connected and labeled regions along with the segmentation. This can help in spatiotemporal tracking, detection of specific objects and real-time video segmentation. A number of preliminary results are shown in Figure~\ref{fig:vidSegResults1}, on the Canoe video sequence from the Change-Detection datasets~\cite{CDW}. The results are obtained by simply changing the distance measure. The second column is obtained by comparing the blue channel of each frame to moderate blue (RGB: 0,0,220) with threshold of 72. Third column is obtained in a similar way by comparing with any reddish pixel of the Canoe (example RGB: 140, 65, 90) with threshold 72. In both cases, distance value must be lower than 72 to be \emph{similar}. Fourth column demonstrates the dual of the criteria for second column: threshold on blue channel but distance value must be higher than 72.

It is a legitimate assumption that the scene content does not change abruptly, for two subsequent frames of a video. Thus, based on the similarity of pixels in a region, a segmented region should have similar mean color in two subsequent frames. This considerably reduces the flickering of colors in a video segmentation. A number of subsequent frames are shown in Figure~\ref{fig:vidSegResults2} for a threshold of 15 on tpsEuclid. Although this is a preliminary application, better segmentation quality may be achieved with different distance measures.

\begin{Figure}
\begin{center}
\includegraphics[width=\linewidth]{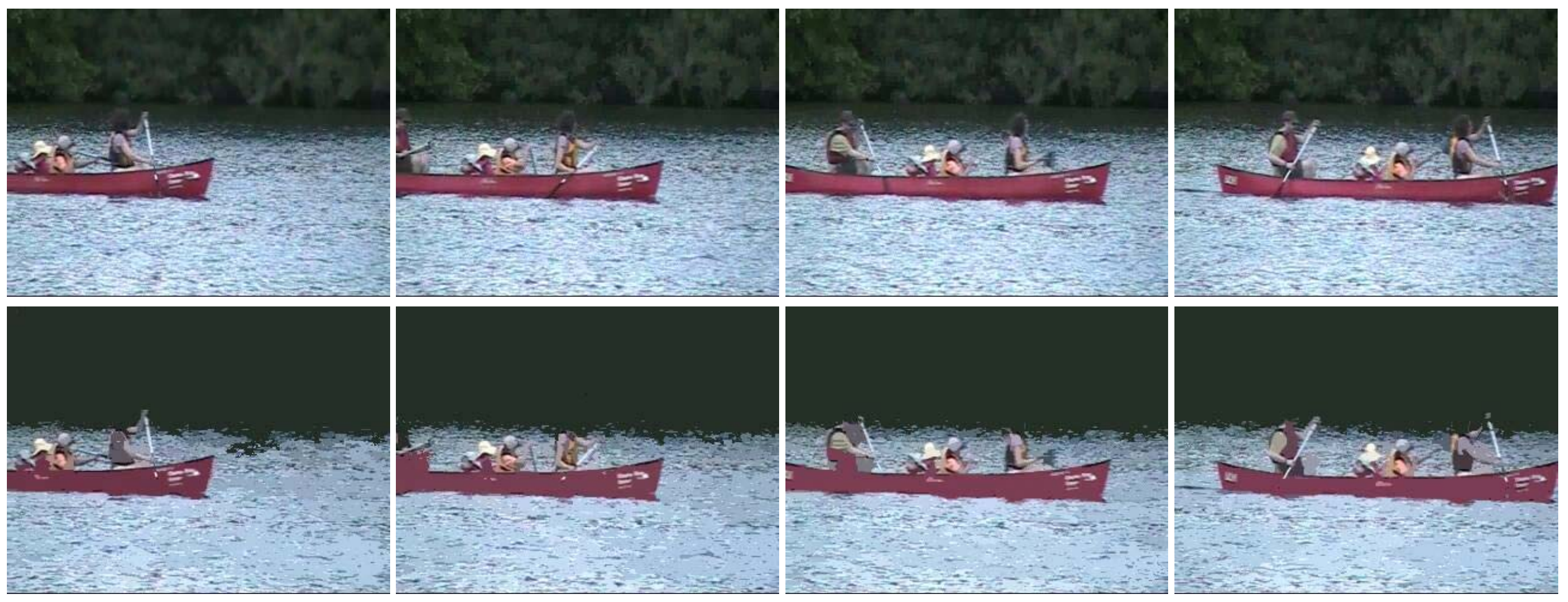}
\end{center}
\vspace{-0.1in}
\captionof{figure}{\small Video segmentation: First row shows four frames 910, 930, 950 and 970, respectively. Second row shows corresponding segmentations.}
\label{fig:vidSegResults2}
\vspace{-0.05in}
\end{Figure}

\section{Conclusion}
\label{sec:conclusion}

In this work, we proposed a novel method for multi-value segmentation based on connected components analysis. Primarily, the method enables us to group similar regions in multi-valued image. The similarity criteria for segmentation can be custom defined. The efficacy of the method has been demonstrated by its use on real-time image and video segmentation. It is worth noticing, that there is no need of any seed or number of clusters for the process of segmentation. However, one of the drawbacks lies in defining the appropriate distance measure for a particular application. The distance measures applied in the work have a high dependency on texture variations. Thus, they fail in high-textured regions and are prone to segment leakage. Thus, future advancements would include searching for a distance measure relatively insensitive to high texture variations.

{\small
\bibliographystyle{ieee}
\bibliography{refs}
}
\end{multicols}
\end{document}